\documentclass{iosart2c}

\usepackage[T1]{fontenc}
\usepackage{times}%
\usepackage[square,sort,comma,numbers]{natbib}
\usepackage{amsmath}
\usepackage{dcolumn}
\usepackage{graphics}
\usepackage[table,dvipsnames]{xcolor}
\usepackage{amssymb}
\usepackage{amsmath} 
\usepackage{comment}
\usepackage{pgfplots}
\usepackage{booktabs}
\usepackage{arydshln}
\usepackage{multirow}
\usepackage{graphicx}
\usepackage{subcaption}
\usepackage{url}

\usepackage{pifont}
\newcommand{\cmark}{\ding{51}}%
\newcommand{\xmark}{\ding{55}}%

\definecolor{ultramarine}{RGB}{0,32,96}
\definecolor{neworange}{RGB}{255,87,51}
\usepackage{amsthm}
\theoremstyle{plain}
\newtheorem{thm}{Theorem} 

\theoremstyle{definition}
\newtheorem{defn}[thm]{Definition} 
\newtheorem{exmp}{Example}[section] 
\usetikzlibrary{fit}

\newcommand{\dquotes}[1]{``#1''}

\newcolumntype{d}[1]{D{.}{.}{#1}}

\definecolor{blue(pigment)}{rgb}{0.2, 0.2, 0.7}

\firstpage{1} \lastpage{5} \volume{1} \pubyear{2020}

\begin{document}
\begin{frontmatter}                           

%
\title{Prioritized Multi-Criteria Federated Learning}

\runningtitle{Prioritized Multi-Criteria Federated Learning}

\author{\fnms{Vito Walter} \snm{Anelli}},
\author{\fnms{Yashar} \snm{Deldjoo}},
\author{\fnms{Tommaso} \snm{Di Noia}}
and
\author{\fnms{Antonio} \snm{Ferrara}\thanks{Authors are listed in alphabetical order. Corresponding author. E-mail: antonio.ferrara@poliba.it.}}
\runningauthor{Anelli, Deldjoo, Di Noia, Ferrara}
\address{Dipartimento di Ingegneria Elettrica e dell'Informazione, Politecnico di Bari, Via E. Orabona 4, 70125, Bari,\\Italy\\
E-mail: firstname.lastname@poliba.it}

\begin{abstract}
In Machine Learning scenarios, privacy is a crucial concern when models have to be trained with private data coming from users of a service, such as a recommender system, a location-based mobile service, a mobile phone text messaging service providing next word prediction, or a face image classification system. 
The main issue is that, often, data are collected, transferred, and processed by third parties. These transactions violate new regulations, such as GDPR. Furthermore, users usually are not willing to share private data such as their visited locations, the text messages they wrote, or the photo they took with a third party.
On the other hand, users appreciate services that work based on their behaviors and preferences.
In order to address these issues, Federated Learning (FL) has been recently proposed as a means to build ML models based on private datasets distributed over a large number of clients, while preventing data leakage. 
A federation of users is asked to train a same global model on their private data, while a central coordinating server receives locally computed updates by clients and aggregate them to obtain a better global model, without the need to use clients' actual data.
In this work, we extend the FL approach by pushing forward the state-of-the-art approaches in the aggregation step of FL, which we deem crucial for building a high-quality global model. Specifically, we propose an approach that takes into account a suite of \textit{client-specific criteria} that constitute the basis for assigning a score to each client based on a priority of criteria defined by the service provider.
Extensive experiments on two publicly available datasets indicate the merits of the proposed approach compared to standard FL baseline.
\end{abstract}


\end{frontmatter}

\section{Introduction}
\label{sec:intro}

The vast amount of data generated by billions of mobile and online IoT devices worldwide holds the promise of significantly improving usability and user experience in intelligent applications. Such large-scale quantity of rich data has created an opportunity to dramatically advance the intelligence of machine learning models by catering powerful deep neural network models. Despite this opportunity, we know that such pervasive devices can capture much data about the user, such as what she does, what she sees, and even where she goes~\cite{DBLP:journals/itpro/MillerVH12}, where most of these data contain sensitive that a user may deem private.

To respond to concerns about the sensitivity of user data in terms of data privacy and security~\cite{DBLP:journals/corr/abs-2005-10322,zerka2020systematic,DBLP:conf/wsdm/DeldjooNM20}, in recent years initiatives have been made by governments to prioritize and improve the security and confidentiality of user data. For instance, in 2018, the General Data Protection Regulation (GDPR) was enforced by the European Union to protect individuals' privacy and data security. These issues and regulations pose a new challenge to traditional AI models where one party is involved in collecting, processing, and transferring all data to other parties. It is easy to foresee the risks and responsibilities involved in storing/processing such sensitive data in the traditional \textit{centralized} AI fashion.

Federated learning is an approach recently proposed by Google~\cite{konevcny2015federated,konevcny2016federated,mcmahan2017communication} to train a global machine learning model from a massive amount of data, which is \textit{distributed} on the client devices such as personal mobile phones and IoT devices. It is noteworthy that FL differs from traditional distributed learning since we assume that training data --- which is supposed to be sensitive --- is maintained on the users' devices they were generated on (e.g., data generated from users' interaction with mobile applications, such as geographical data in location-aware applications~\cite{RDACP16}).
Therefore, we have to deal with data that is quantitatively unbalanced and differently distributed over devices, i.e., each device data is not a representative sample of the overall distribution. Instead, in a traditional distributed setting, data has to be collected in a centralized location and then evenly spread over proprietary computing nodes.  As a matter of fact, with FL, we leverage users' computing power for training a shared ML model while preserving privacy, by actually decoupling the ability to learn a ML model from the need to store private data centrally.

In principle, a FL model is able to deal with fundamental issues related to privacy, ownership and locality of data~\cite{DBLP:journals/corr/abs-1902-01046}.  In~\cite{mcmahan2017communication}, authors introduced the \textit{FederatedAveraging} (FedAvg) algorithm, which combines local stochastic gradient descent on each client via a central server that performs model aggregation by averaging the values of local parameters. To ensure that the developments made in FL scenarios uphold to real-world assumptions, in~\cite{caldas2018leaf}  the authors introduced LEAF, a modular benchmarking framework supplying developers/researchers with an abundant number of resources including open-source federated datasets, an evaluation framework, and several reference implementations.

Despite its potentially disruptive contribution, we argue that FedAvg has several significant shortcomings. First, the aggregation operation in FedAvg assigns the contribution of each agent proportional to each client’s local dataset size. This simplifying assumption ignores a wealth of other qualitative measures that can be impactful for training an effective global model. Examples of such measures include the number of sample classes held by each agent, the divergence of each computed local model from the global model — which may be critical for convergence~\cite{sahu2018convergence} —, estimations about the agent computing and connection capabilities and finally the amount of client’s honesty and trustworthiness.

While FedAvg only uses limited knowledge about local data, we argue that the integration of the aforementioned qualitative measures and the expert's domain knowledge is fundamental for increasing the global model's quality.
As a toy example, let us consider a federated scenario with just two users Alice and Bob. The photos in users' mobile phones are the training samples of a ML model for classifying clothes. Let us suppose that Alice holds tens of very similar photos with the same outfit, and most of them are blurred. Thus these images do not carry much information. Instead, Bob holds a smaller number of well-labeled photos, but high-quality, and with a lot of different clothes. We argue that in such a situation, the weight of Bob's contribution to the ML model should be higher or comparable to Alice's.

The work at hand considerably extends the FedAvg approach~\cite{mcmahan2017communication} by building on two main assumptions:

\begin{itemize}
    \item we can substantially improve the quality of the global model by incorporating \textit{a set of criteria} measuring some quality properties of the clients, and properly assigning the contribution of individual update in the final model based on these criteria;
    
    \item the introduced criteria can be combined by using a \textit{score function} able to \dquotes{summarize} them; toward this goal, we assert about the potential benefits of using a \textit{prioritized multi-criteria} score function over the identified set of criteria.
\end{itemize}

\noindent Moreover, the FL system proposed in this work moves one step forward from our previous effort in~\cite{DBLP:conf/aiia/AnelliDNF19} along the following directions:
(i) a formal notion of \textit{criterion} is proposed and a formal classification for possible criteria is summarized (cf.~Section~\ref{subsec:Iden_LocalCrit}), (ii) a mathematical formalization for score functions is described (cf. Section~\ref{subsec:Iden_ScoreFun}), (iii) the system is evaluated through more comprehensive experiments, in particular, we compare our approach against FedAvg both on MNIST dataset (with IID and Non-IID distribution), used in~\cite{mcmahan2017communication}, and on CelebA dataset, where a classification task is performed by also injecting more explicit information about the quality of the local dataset and local model.

The remainder of the paper is structured as follows. In Section~\ref{sec:RL} we review some related work, while Section~\ref{sec:FL_BKG} is devoted to introduce the standard FL model and its key concepts. In Section~\ref{sec:FL_proposed} we provide a formal description of the proposed FL approach and we formalize both the concepts of local criteria and score function, while providing some general concepts for their choice. Section~\ref{sec:expe} details the experimental setup, as well as the datasets, the tasks and the metrics we consider for validating our approach and studying its variables. Section~\ref{subsec:eval_FL_proposed} presents results and discussion. Finally, Section~\ref{sec:conlusion} concludes the paper and discusses future perspectives.

\section{Related Work}
\label{sec:RL}

\begin{figure}
    \centering
    \includegraphics[width=\columnwidth]{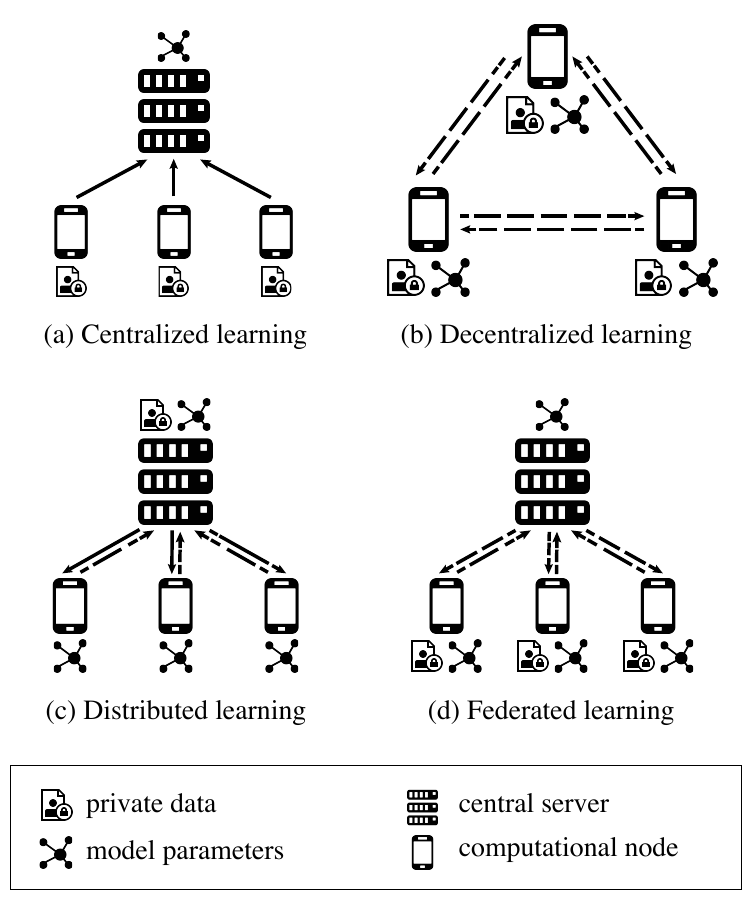}
    \caption{Information flow over the network in four ML architectures. Solid lines represent training data flow, dash lines represent model parameters flow.}
    \label{fig:learning_par}
\end{figure}

Nowadays, any AI application requires capturing a considerable amount of users’ sensitive data about their routine activities. Moreover, users may be unaware of how their data are collected, processed, and transferred between parties~\cite{DBLP:journals/tist/YangLCT19}.
Federated learning aims to meet the privacy ML shortcomings by horizontally distributing the training of the model over user devices, thus making clients exploit private data without sharing them~\cite{mcmahan2017communication}.

Federated Learning ties in with other privacy-preserving ML approaches, aiming to equip ML algorithms with defense measures for protecting user privacy and data security. Beyond the recent FL paradigm, the field of privacy preservation in ML has been thoroughly investigated through the lens of (i) Secure Multi-Party Computation~\cite{DBLP:conf/focs/Yao86}, (ii) Homomorphic Encryption~\cite{DBLP:conf/stoc/Gentry09}, and (iii) Differential Privacy~\cite{DBLP:journals/fttcs/DworkR14, DBLP:journals/prl/OnetoRA17}, which enable different types of defense against privacy threats in ML.

Figure~\ref{fig:learning_par} highlights the differences between different learning approaches, including Federated Learning. Four architectures are compared, focusing on the information exchanged over the network: the solid lines represent training data flow, the dashed lines represent model parameters flow. In detail, we distinguish between the following learning paradigms from a distribution point-of-view:

\begin{itemize}
    \item \textbf{centralized learning}, where a central server is in charge of data collection, data aggregation and model training (the data flows from clients to the server); 
    \item \textbf{decentralized learning}, also known as a peer-to-peer (P2P) architecture, in which clients share resources directly with each other without the need of a centralized control, therefore the model parameters are exchanged between clients, as shown by dashed line in Figure~\ref{fig:learning_par};
    \item \textbf{distributed learning}, in which a centralized server collects user data and redistributes them to other computational nodes, in order to exploit their computational resources;
    \item \textbf{federated learning}, the approach proposed by Google in the context of machine learning, in which a central server still makes use of computational resources of other clients, without the need to collect their data: users' data remain on their clients, and only model parameters are exchanged between the server and the clients.
   
\end{itemize}

It is interesting to notice how FL was originally conceived as a cross-device architecture involving users' mobile devices and edge device applications, e.g., for Google's Gboard mobile keyboard~\cite{hard2018federated}. Nonetheless, FL also extended towards cross-silo architectures~\cite{kairouz2019advances}, for multiple organizations willing to train a model in inherently sensitive domains (e.g., financial or medical). For example in healthcare, regulations prohibit medical institutions from sharing medical data (e.g., to improve medical imaging performance~\cite{vizitiu2019towards}); however, FL makes centralization unnecessary, by allowing data to remain isolated in the individual organizations while still improving medical AI models~\cite{zerka2020systematic}.

While handling users' privacy concerns, FL faces challenges such as  communication costs, unbalanced data distribution and device reliability and security issues (e.g., model poisoning~\cite{DBLP:journals/corr/abs-1807-00459}, indirect information leakage~\cite{DBLP:journals/tifs/PhongAHWM18}, and Byzantine adversaries~\cite{DBLP:journals/pomacs/SuX19}). Despite its original formulation, in \cite{DBLP:journals/tist/YangLCT19} federated learning concept is extended to a more comprehensive idea of privacy-preserving decentralized collaborative ML techniques, both for horizontal federations (where different datasets share the same feature space but are different in training samples) and vertical federations (where different datasets share the  training samples but they differ in feature space).

In literature, a growing number of works focus on some modifications of FedAvg algorithm, e.g., adding a regularization term in the local objective functions for controlling the convergence of the network with statistical heterogeneity of data~\cite{sahu2018convergence}, changing the optimization algorithm~\cite{DBLP:journals/corr/abs-2003-00295}. Some other works focus on some adaptations for controlling the averaging operation on a per-layer basis~\cite{DBLP:journals/corr/abs-2002-06440} or for obtaining more 
personalization on the local models~\cite{DBLP:journals/corr/abs-2002-07948}. Finally, the work in~\cite{DBLP:journals/access/YeYPH20} studies a protocol for selecting devices based on their dataset quality without any information disclosure.

The approach we propose is configured as a formal tool for incorporating any of the above-mentioned objectives (for example, facing the statistical challenges), when these could be obtained by pushing up or down the contribution of each client, based on some requirements. Such mechanism is completely incorporated in the weight of each client in the global aggregation. To the best of our knowledge, this is the only work attempting to modify the value of the weight of clients in the aggregation step.

\section{Background Technology}
\label{sec:FL_BKG}

In recent years, Federated Learning (FL) have been proposed by Google~\cite{mcmahan2017communication, konevcny2016federated, konevcny2016federated2} as a privacy-preserving architecture for training a ML model, by using private data from a large set of users who will never share them with other parties in the system. 
The motivation behind FL is mainly efficiency and privacy~\cite{bagdasaryan2018backdoor,mcmahan2017communication}.
Indeed, since users' data never leaves the devices, FL models can be trained on private and sensitive data, e.g., the history of user typed messages, which can be considerably different from publicly accessible datasets.

\begin{table}[!tbp]
\setlength\tabcolsep{1.9pt}
\caption{Definition of variables used in this work.}
\label{tbl:abbrv}
\centering
\begin{tabular}{cc}
\toprule
\multicolumn{1}{c}{\textbf{Notation}} & \multicolumn{1}{c}{\textbf{Description}} \\ \toprule 
$S$ & central server \\
$\mathcal{A}$ & set of users of the federation \\
$\mathcal{A}^- \subseteq \mathcal{A}$ & subset of users taking part to a round of communication \\
$a \in \mathcal{A}$ & a user \\
$\mathcal{D}_a$ & user $a$'s training dataset \\
$\Theta$ & ML model parameters \\
$G$ & loss function \\
$\alpha$ & learning rate \\
$w_a$ & aggregation weight associated to user $a$ \\
$f_m$ & score function working on $m$ criteria \\
$\mathcal{C}$ & set of measurable criteria over clients \\
$C_i \in \mathcal{C}$ & the $i$-th criterion \\
$c_{i,a}$ & degree of satisfaction of the $i$-th criterion for client $a$ \\
\bottomrule
\end{tabular}
\end{table}

Although FL principles can be applied to many machine learning tasks, throughout this paper we consider a fundamental problem in ML, classification task. Assume a training dataset $\mathcal{D}$ composed of $n$ pairs $\{(x_i, y_i) \in \mathcal{X} \times \mathcal{Y} \mid i=1, \dots, n\}$ whose elements are input samples $x_i \in \mathcal{X}$ and corresponding class labels $y_i \in \mathcal{Y}$. It is presumed that this dataset is i.i.d. drawn from an unknown distribution $\mathcal{P}$. The problem of classification is often expressed as finding a function $f_\Theta: \mathcal{X} \rightarrow \mathcal{Y}$ that can approximate the class labels around the input samples, where $\Theta \in \mathbb{R}^L$ represents the model parameters and $L$ is the dimensionality of the parameters space. Finding the model parameters $\Theta$ is achieved by solving an empirical risk minimization (ERM) problem of the form
\begin{equation}
   \Theta^* = \min_{\Theta} \sum_{(x_i, y_i) \in \mathcal{D}} G(f(x_i;\Theta),y_i)
\end{equation}
where $G(\cdot)$ is the empirical risk function (a.k.a. loss function).

In the federated learning (FL) setting, there are two main players:
\begin{itemize}
        \item \textbf{Agents.} We denote with $\mathcal{A}$ the set of agents who participate to the federation process. Each agent $a \in \mathcal{A}$ owns its local data $\mathcal{D}_a$ (referred to as a \textit{shards} or \textit{local datasets}) and never shares it with other parties of the system. Formally, $\mathcal{D}_a = \{(x_i, y_i) \in \mathcal{X} \times \mathcal{Y} \mid i=1, \dots, n_a\}$, where $n_a$ denotes the number of data points that the agent $a$ holds.\footnote{An agent represents a computational node of the system, usually corresponding to the private device of an user, which has the role of a client in the architecture. For this reason, throughout the paper, we will use the three terms \textit{clients}, \textit{agents} and \textit{users} interchangeably.}
        \item \textbf{Central Server ($S$).} The central server $S$ represents the main component in a FL setting. The server does not have access to local shards. Its goal is to train the classifier $f_\Theta$ characterized by the global parameters vector $\Theta \in R^L$ by distributed training, collecting local parameters, and aggregating them over the number of agents involved in the federation.
\end{itemize}

In the following, we briefly review the principles of Federated Learning by means of a formal definition, based on the scenario presented so far.

\begin{defn}[Federated Learning]
In classical FL setting, at each round of communication (a time step in the training process), the server $S$ selects a fraction of clients $\mathcal{A}^- \subseteq \mathcal{A}$ from the federation and shares with them the current global model $\Theta$. Every agent $a \in \mathcal{A}^-$ uses it own private shard $\mathcal{D}_a$ in order to minimize an \textit{empirical local loss} $G_{\mathcal{D}_a}(\Theta)$. For this, the agent $a$ takes the global parameter $\Theta$ as the initial point and runs an iterative method --- such as Stochastic Gradient Descent (SGD) --- for a fixed number of epochs $E$ and batch size $B$ which results in a variation of the local gradient $\nabla G_{\mathcal{D}_a}(\Theta)$ for each user and then it communicates the \textit{gradient} to the server $S$. Once $S$ has collected all the gradients for each $a \in \mathcal{A}^-$, at the end of the round of communication, it updates the new global model $\Theta$ as:
\begin{equation}
\label{eq:aggreg}
\Theta \leftarrow \Theta - \alpha \sum_{a \in \mathcal{A}^-} w_a \nabla G_{\mathcal{D}_a}(\Theta),
\end{equation}
\noindent in which  $w_a \in [0,1]$ is the weight associated with agent $a$ and $\sum_{a \in \mathcal{A}^-} w_a = 1$.
It is worth noticing that such formulation builds on the assumption that $\sum_{a \in \mathcal{A}} w_a \nabla G_{\mathcal{D}_a}(\Theta) = \nabla G_{\mathcal{D}}(\Theta)$, where $\mathcal{D}=\bigcup_{a \in \mathcal{A}} \mathcal{D}_a$ represents an ideal centralized dataset.
\hfill\(\Box\)
\end{defn}

There exists an alternative for this formulation, in which the client $a$ updates the received global model $\Theta$ and calculates a local version $\Theta_a$ according to Eq.~\ref{eq:alt} and then it communicates the updated local model $\Theta_a$ (instead of the local gradients) to $S$:
\begin{equation}
\label{eq:alt}
    \Theta_a \leftarrow \Theta - \alpha \nabla G_{\mathcal{D}_a}(\Theta).
\end{equation}
Once the end of communication rounds has been reached, then $S$ computes the new global model $\Theta$ as:
\begin{equation}
\label{eq:alt_global}
    \Theta \leftarrow \sum_{a \in \mathcal{A}^-} w_a \Theta_a.
\end{equation}

A well-known implementation of FL is \textit{FedAvg}~\cite{mcmahan2017communication},  in which the authors use the FL procedure to train a \textit{deep learning} model using SGD for a classification task.

\section{Prioritized Multi-Criteria Federated Learning}
\label{sec:FL_proposed}
The FL principles discussed in Section~\ref{sec:FL_BKG} prevent the system from collecting information about users, giving rise to a natural trade-off between users' data privacy and performance of the system. Our assumption is that revealing some \textit{non-sensitive client-related information} (shared by each client $a$) and integrating this knowledge in the global aggregation step represented by Eq.~\ref{eq:alt_global} (performed by the server $S$) could lead to learning more effective federated model without harming users' privacy. For instance, such non-sensitive data may carry useful information about specific domain or some marketing objectives that can be leveraged to build more \textit{in-domain} strategies or increase the system \textit{profitability}.

The proposed solution envisioned in this work is a client-aware FL strategy based on the following elements:

\begin{itemize}
    \item we introduce the notion of \textit{criterion}, which measures a specific property about users and their data, and we propose a formal classification for them;
    
    \item we propose the use of a \textit{score function} to \dquotes{summarize} the criterion measurements and compute a score for each client of the federation to be encoded in the aggregation step of each round of communication;
    
    \item we explore the benefits of a \textit{prioritized multi-criteria score function} over the identified set of criteria;
\end{itemize}

In the following, we introduce the fundamentals and the formalism behind our approach.

\subsection{Fundamentals of the proposed approach}
\label{sec:fund-approach}
Let us assume $\mathcal{C}=\{C_1,...,C_m\}$ denote a set of measurable properties (a.k.a. criteria) characterizing the local agent $a$ or local data $\mathcal{D}_a$. We use the variable $c_{i,a} \in [0,1]$ to denote, for each agent $a$, the degree of satisfaction of the criterion $C_i$ in a specific round of communication.
We assume the clients compute these values and that they communicate with the server the correct model updates and the correct property measurements (honest clients). 
Hence, at the end of the communication rounds, the server $S$ is in charge of collecting not only the model update $\Theta_a$, but also the $m$-tuple $\mathbf{c}_a = (c_{1,a}, \dots, c_{m,a})$.
To ensure the same scale for each criterion, we assume that the measure $c_{i,a}$ of the $i$-th criterion on device $a$ is a real value in the interval $[0, 1]$ (with $0$ meaning unfulfillment and $1$ meaning total adherence to the criterion).
Additionally, the measurements of $C_i$ over all devices in $\mathcal{A}^-$ are supposed to be normalized, i.e., $\sum_{a \in \mathcal{A}^-} c_{i,a} = 1$.

The main idea of the work at hand is that we can encode the knowledge about clients in the weights $w_a$ used for aggregating client contributions in Eq.~\ref{eq:aggreg}. Based on that, $S$ can compute the weight $w_a$ for agent $a$ according to the following equation:
\begin{equation}
\label{eq:score}
    w_a = \frac{f_m(\mathbf{c}_a)}{Z} = \frac{f_m(c_{1,a},...,c_{m,a})}{Z},
\end{equation}
\noindent where $f_m: [0,1]^m \rightarrow \mathbb{R}$ is a score function over the $m$-tuple of properties (criteria), which evaluates agent $a$'s contribution based on the fulfillment of such criteria. Finally, $Z$ is a normalization factor introduced to ensure that $\sum_{a \in \mathcal{A}^-}w_a=1$ and $w_a \in [0,1]$; therefore,  $Z=\sum_{a \in \mathcal{A}^-} f_m(\mathbf{c}_a)$.

\begin{exmp}
Let us go back to the toy example we introduced in Section~\ref{sec:intro}, with $\mathcal{A}^- = \{\textit{Alice}, \textit{Bob}\}$. Let us consider the set $\mathcal{C}$ of criteria describing the two clients of the federation, i.e., specific qualities related to their local devices, produced local models and local data; for example we may consider \textit{\textbf{D}ataset \textbf{S}ize}, \textit{\textbf{C}lothes \textbf{D}iversity}, and \textit{\textbf{I}mage \textbf{S}harpness} thus having  $\mathcal{C}=\{DS, CD, IS\}$.
Assume that Alice has received evaluations $c_{\textit{DS},\text{Alice}} = 0.9$, $c_{\textit{CD},\text{Alice}} = 0.2$, $c_{\textit{IS},\text{Alice}} = 0.4$, while Bob has obtained $c_{\textit{DS},\text{Bob}} = 0.1$, $c_{\textit{CD},\text{Bob}} = 0.8$, $c_{\textit{IS},\text{Bob}} = 0.5$. Based on Eq.~\ref{eq:score}, the overall contribution of Alice and Bob will be $f_3(0.9, 0.2, 0.4)$ and $f_3(0.1, 0.8, 0.5)$, respectively, both divided by $Z=f_3(0.9, 0.2, 0.4)+f_3(0.1, 0.8, 0.5)$ to obtain the weights $w_\text{Alice}$ and $w_\text{Bob}$. To get an idea, if we consider the score function $f_3$ to be a basic mean operation, the final values for $w_\text{Alice}$ and $w_\text{Bob}$ would be: $w_\text{Alice} = \frac{1.5}{1.5+1.4}=0.52$, $w_\text{Bob} = \frac{1.4}{1.5+1.4}=0.48$.
\hfill\(\Box\)
\end{exmp}

In the following, we detail the main dimensions of the study related to the proposed approach, which include:

\begin{itemize}
    \item identification of the set $\mathcal{C}$ of criteria. These criteria can be related to users (e.g., gender), clients (e.g., number of samples), or local models (e.g., the local model highly diverges from the centralized model). Besides, the criteria can express a boolean fulfillment of the requirement (e.g., Female: True or False), or a quantitative estimation (e.g., the value of the divergence of the model);
    \item identification of a score function able to \dquotes{summarize} the computed measurements in a score value representing each agent.
\end{itemize}

To properly study each of the presented dimensions, in the following we start discussing about the identification of criteria, then we go along the study of score functions and, in particular, priority-based score functions.

\subsection{Identification of local criteria}\label{subsec:Iden_LocalCrit}

In the original FedAvg formulation, the server performs aggregation to compute $w_a$, without knowing any information about participating clients, except for a pure quantitative measure about local dataset size. 

However, a federated setting has to face some key issues~\cite{mcmahan2017communication} related to communication, connectivity and statistics.
Among them, if we focus on the statistical aspects about the data, we notice that training data --- which is typically the result of the real user usage of the device --- represent a non-IID distribution of the training data over the large number of clients. Moreover, the amount of data is also unbalanced across the clients.
Such characteristic may affect the accuracy and the efficiency of the resulting aggregated model~\cite{noniid}.
In this scenario, a service provider may be interested in defining some statistical criteria such that the rounds of communication needed to reach the desired target accuracy are minimized. This result can be accomplished by enhancing the contribution of clients fulfilling the requirements expressed by the defined criteria.

For instance, useful information could be related, in a classification problem, to the number of classes covered by a local dataset.
Moreover, a domain expert could ask users to measure their adherence to some other target properties (e.g. their nationality, gender, age, job, behavioral characteristics, etc.), in order to build a global model emphasizing the contribution of some classes of users; in this way, the domain expert may, in principle,  build a model favoring some targeted commercial purposes.

We identified four classes of local criteria, each of them related to different aspects of the local dataset $\mathcal{D}_a$, the local device or the user $a$:

\begin{itemize}
    \item criteria describing the quality of the local dataset $\mathcal{D}_a$ (e.g., dataset size, diversity of training samples, etc.);
    \item criteria describing the quality of the produced local model $\nabla G_{\mathcal{D}_a}$ (e.g., ...);
    \item criteria describing the trustworthiness of device $a$;
    \item criteria capturing the fulfillment of user $a$ with respect to commercial targets (e.g., gender, job, nationality, etc.).
\end{itemize}

Although we can identify some classes of criteria, the choice for each particular criterion still remain a strictly task- and domain-dependent activity. We provide some insights about how to define such criteria: first, as a rule of thumb, one should start from the objective, e.g., obtaining faster convergence, overshadow unreliable updates, specialize the model with respect to some categories of users and then choose characteristics which can improve the model towards that initial objective. Next, the system designer could play with the identified criteria to test their effectiveness, e.g. with some previous centralized data, or with some synthetic data, or with some validation rounds of federated training. Finally, the empirical evaluations performed in this work suggest that choosing criteria that lead to higher variance in the score obtained across clients results in a better final model. For instance, the criterion dataset size is not an appropriate criterion, if all the clients have the similar number of samples in their private datasets.

For this reason, in this work we do not focus on a particular choice for them, but rather on presenting a formal approach about dealing with them. Moreover, we show in the experimental section --- for illustrative purposes only --- some examples of how they can be chosen with respect to the task.

\subsection{Identification of a score function}\label{subsec:Iden_ScoreFun}

A crucial aspect of this work is related to the identification of a score function $f_m$ able to \dquotes{summarize} the values $c_{i,a}$ for each criterion $C_i \in \mathcal{C}, i \in [1, \dots, m]$ in order to obtain the value $w_a$ as described in Eq.~\ref{eq:score}. The score has to be computed for each client separately, so in the following we will focus on the computation of such value for one client $a \in \mathcal{A}$.

Over the years, a wide range of aggregation functions have been proposed in the field of information retrieval (IR)~\cite{pasi}. Just to mention the most relevant ones, we can refer to the weighted averaging operator, as well as to the ordered weighted averaging (OWA) models~\cite{DBLP:journals/ijis/Yager96,DBLP:journals/tsmc/Yager88} --- which extend the binary logic of \textit{AND} and \textit{OR} operators, allowing representation of intermediate quantifiers ---, to the Choquet-based models~\cite{AIF195451310,Grabisch2000,GRABISCH1996445} --- which are able to interpret positive and negative interactions between criteria ---, and finally to the priority-based models~\cite{dacosta}. 

Although we have many opportunities, we focus on the priority-based score function proposed in~\cite{dacosta}. Such function can be formulated with a compact notation as:
\begin{equation}
\label{eq:pasi}
f_m(\mathbf{c}) = \sum_{i=1}^m \prod_{j=1}^i c_j,
\end{equation}
where $m$ represents the number of values the function is applied to. This function holds the monotonic property
\begin{multline}
        f_m(c_1, \dots, c_i, \dots, c_m) \leq f_m(c_1, \dots, c_i', \dots, c_m) \\ \forall c_i \leq c_i', \quad i=1,\dots, m.
\end{multline}
and, at the same time, we have $f_m(\mathbf{0}) = 0$ and $f_m(\mathbf{1}) = m$, i.e., the final score is zero when all values are 0, and it is maximum when all values are 1.
One of the most interesting properties of this function is that:
\begin{multline}
\label{eq:anni}
    f_m(c_1,\dots,c_{j-1},0,c_{j+1},\dots,c_m) \\ = f_{j-1}(c_1,\dots,c_{j-1}), \quad
    \forall j \in {1, \dots, m}.
\end{multline}
\noindent This means that the lack of fulfillment of a higher  criterion in the list cannot be compensated with the fulfillment of a lower one~\cite{pasi}. If we adopt a priority order for the criteria in $\mathbf{c}$ from the higher to the lower, it follows that in case a higher criterion is equal to $0$ then it cannot be compensated by criteria with a lower priority. Interestingly, we also have that $f_m(0,c_2,\dots,c_m)=0$. 

The main aim of such function, in a multi-criteria scenario, is to use a priority order over the involved criteria in order to assign each agent $a$ a score based on the measurements $c_{i,a}$, for each $C_i \in \mathcal{C}, i \in [1, \dots, m]$, in a priority-aware fashion. With this function, when a criterion is met, the more it is fulfilled the more the subsequent criteria will be taken into account; analogously, the less a criterion is fulfilled the less the other criteria will be considered and summed up. Moreover, based on Eq.~\ref{eq:anni}, when a criterion is not satisfied at all, all the subsequent criteria will not be considered.

The properties presented so far make this function our main choice for our approach. As an example, we may consider the case where the domain expert may wish to consider extremely important the age of a user rather than its dataset size, so that even a large local dataset would be penalized if the user age criteria is not satisfied.

In the following, we assume, for the sake of clarity of notation, that the set $\mathcal{C}$ of criteria can be written in the form $\mathcal{C} = \{C_1, \dots, C_m\}$ where each index represents an identifier for the corresponding criterion, or in the form $\mathcal{C} = \{C_{(1)}, \dots, C_{(m)}\}$, where each index represent the priority of the criterion, from the highest to the lowest one. Analogously, we distinguish between $c_{i,a}$ and $c_{(i),a}$, which represent the measurement on device $a$ of the criterion $C_i$ and the $i$-th important criterion, respectively.

\begin{exmp}
(Cont'd) We carry on with the example about Alice and Bob, where $c_{\textit{DS},\text{Alice}} = 0.9$, $c_{\textit{CD},\text{Alice}} = 0.2$, $c_{\textit{IS},\text{Alice}} = 0.4$, while Bob has obtained $c_{\textit{DS},\text{Bob}} = 0.1$, $c_{\textit{CD},\text{Bob}} = 0.8$, $c_{\textit{IS},\text{Bob}} = 0.5$. We have already shown that with a simple mean function their final scores were $w_\text{Alice} = \frac{1.5}{1.5+1.4}=0.52$, $w_\text{Bob} = \frac{1.4}{1.5+1.4}=0.48$, with Alice obtaining higher score than Bob. Let us consider now the prioritized score function and let us see how score changes based on priority. Suppose that $C_{(1)}= DS$, $C_{(2)}= CD$, $C_{(3)}= IS$.
Based on Eq.~\ref{eq:pasi},
\begin{multline*}
    f_3(\mathbf{c}_\text{Alice}) = f_3(c_{(1),\text{Alice}}, c_{(2),\text{Alice}}, c_{(3),\text{Alice}}) \\
    = f_3(c_{\textit{DS},\text{Alice}}, c_{\textit{CD},\text{Alice}}, c_{\textit{IS},\text{Alice}}) \\
    = f_3(0.9, 0.2, 0.4) 
    \\ = 0.9 + (0.9 \cdot 0.2) + (0.9 \cdot 0.2 \cdot 0.4) = 1.152,
\end{multline*}
\noindent while $f_3(\mathbf{c}_\text{Bob}) = 0.22$. If we change the priority order to be $C_{(1)}=$ image sharpness, $C_{(2)}=$ clothes diversity, $C_{(3)}=$ dataset size, we would then obtain:
\begin{multline*}
    f_3(\mathbf{c}_\text{Alice}) = f_3(c_{(1),\text{Alice}}, c_{(2),\text{Alice}}, c_{(3),\text{Alice}}) \\
    = f_3(c_{\textit{IS},\text{Alice}}, c_{\textit{CD},\text{Alice}}, c_{\textit{DS},\text{Alice}})
    \\ = f_3(0.4, 0.2, 0.9) \\
    = 0.4 + (0.4 \cdot 0.2) + (0.4 \cdot 0.2 \cdot 0.9) = 0.552,
\end{multline*}
\noindent while $f_3(\mathbf{c}_\text{Bob}) = 0.94$. We see that with the second configuration, the score for Alice is lower than the previous one since the most important criterion here is worse fulfilled, conversely for Bob. \hfill\(\Box\)
\end{exmp}

\section{Experimental Setup}
\label{sec:expe}

In this section we describe the experimental setup --- datasets, tasks and identified local criteria --- used to validate the  performance of the proposed FL system. In detail, we validate our approach on the following datasets and tasks:

\begin{itemize}
    \item \textbf{MNIST}~\cite{lecun1998gradient}: handwritten digits; we perform a 10-class classification for recognition of digits; \item \textbf{CelebA}~\cite{liu2015faceattributes}: face images of celebrities coming with 40 different attributes under varying poses and backgrounds; we perform a binary prediction between smiling and non-smiling faces.
\end{itemize}

\noindent For each dataset we consider different data distributions, as well as a specific model. Moreover, for each of them we define different criteria based on possible useful information we may extract.

\subsection{MNIST experiments}\label{subsec:MNIST_exp}
We run a first cluster of experiments on MNIST dataset~\cite{lecun1998gradient} for the digit recognition task. This dataset contains 60,000 examples of 10 classes of handwritten digits (plus a test set of 10,000 examples) by 500 writers. The samples are available in the form of 28x28 pixel black and white images.

We use this dataset in a completely user-agnostic way. Therefore, we created the federated dataset by following the two partitioning ways used in~\cite{mcmahan2017communication}, in order to simulate both a IID distribution and a Non-IID distribution of data over the different clients.

\paragraph{Model.} For the sake of comparability, we built the same classifier described in~\cite{mcmahan2017communication}, i.e., a CNN with two convolution layers with 5x5 filters --- the first layer with 32 channels, the second with 64, each followed with a 2x2 max pooling layer ---, a fully connected layer with 512 units and ReLu activation, and a final softmax output layer with 10 neurons, for a total of 1,663,370 total parameters.

\paragraph{Local criteria.} For this experimental setting, we aim at both reducing the number of rounds of communication necessary to reach a target accuracy and making the global model not diverging towards local specializations and overfittings. To this aim, we extend the pure quantitative criterion in FedAvg~\cite{mcmahan2017communication}  --- dataset size ---  by leveraging two new criteria\footnote{Please note that we are not stating that the proposed ones are the only possible criteria. We present them just to show how the introduction of new information may lead to a better final model.}.

The first criterion we consider is the one already used by FedAvg~\cite{mcmahan2017communication}, namely the local dataset size (\textbf{DS}), given by $c_{1,a}=|\mathcal{D}_a|/|\cup_{i \in \mathcal{A}^-} \mathcal{D}_i|$. This criterion is a \textit{pure quantitative measure} about the local data, which will serve both as a baseline in empirical validation of the results (i.e., when used in isolation) and as part of the entire identified set of criteria in the developed FL system (i.e., when used in a group).

The second considered criterion is the \textit{diversity of labels} (\textbf{LD}) in each local dataset, measuring the diversity of each local dataset in terms of class labels. We assert this criterion to be important since it can provide a clue on how much each device can be useful for learning to predict different labels. To quantify this criterion we use $c_{2,a}=\delta(\mathcal{D}_a)/\sum_{i \in \mathcal{A}^-}\delta(\mathcal{D}_i)$ where $\delta$ measures the number of different labels (classes) present over the samples of that dataset.

With respect to the third criterion, our aim is to reduce the negative effects of a non-IID distribution. Indeed, with non-IID distributions --- and this is the case of our dataset --- model performance dramatically gets worse~\cite{noniid}. Moreover, a large number of local training epochs may lead each device to move further away from the initial global model, towards the opposite of the global objective~\cite{sahu2018convergence}. Therefore, a possible solution inspired by~\cite{sahu2018convergence} is to limit these negative effects, by penalizing higher divergences and highlighting local models that are not very far from the received global model. We evaluate the local model divergence (\textbf{MW}) as $c_{3,a} = \varphi_a / \sum_{i \in \mathcal{A}^-}\varphi_i$ where $\varphi_i = \frac{1}{\sqrt{||\Theta-\Theta_a||_2+1}}$.

\subsection{CelebA experiments}
\label{subsec:CelebA_exp}
CelebFaces Attributes Dataset (CelebA)~\cite{liu2015faceattributes} is a large-scale dataset with 202,599 face RGB images of 10,177 celebrities. Each image comes with 40 binary attributes and differs from the other ones for celebrity pose and background. The task is a binary classification between smiling and non-smiling people, which is an information included within the 40 attributes. We chose this task since the smiling attribute has a good balance in the whole dataset between positive and negative outcomes.

We use this dataset in a completely user-aware way. Indeed, we suppose that each celebrity holds her own photos in her mobile phone gallery. This represent a realistic set of local datasets, which could have been generated from the personal device usage. This distribution is inherently Non-IID, therefore it is a representative scenario for a federated setting.

We used a subsampled version of the dataset (50\% of images). Then, we removed users with less than 5 photos. For each user, we split her private dataset with random hold-out method with a ratio of 80/20. Finally, images have been resized to 64x64 pixels.

\paragraph{Model.} For the CelebA experiments, we built a CNN binary classifier with two convolution layers with 3x3 filters --- the first layer with 32 channels, the second with 64, each followed with a 2x2 max pooling layer ---, a fully connected layer with 512 units and ReLu activation, and a final softmax output layer with 1 neuron, for a total of 8,409,025 total parameters.

\paragraph{Local criteria.} Also in these experiments, our aim is to reduce the number of rounds of communication needed to reach a target accuracy. Three criteria have been used to reach such objective. 

Also in this case, we keep the dataset size criterion (\textbf{DS}). Therefore, $c_{1,a}=|\mathcal{D}_a|/|\cup_{a \in \mathcal{A}^-} \mathcal{D}_i|$. It will serve both as a baseline in empirical validation of the results and as part of the entire identified set of criteria in the developed FL system.

Similarly to what has been done with the MNIST dataset, we want to consider how distributed are the labels of the local samples. In fact, since we are dealing with a binary classification task, we have at most two different classes in each dataset. For this reason, we consider a measure of class balance (\textbf{CB}), i.e. how they are similar in number. To this aim, we define a function $\zeta$, which measure the ratio between the number of samples of the two classes, namely:

\begin{equation}
    \zeta(\mathcal{D}_a) = \frac{\min\{\text{\# pos.}, \text{\# neg.}\}}{\max\{\text{\# pos.}, \text{\# neg.}\}}
\end{equation}

\noindent Then, the second criterion has been defined as $c_{2,a} = \zeta(\mathcal{D}_a)/\sum_{i \in \mathcal{A}^-} \zeta(\mathcal{D}_i)$.

As mentioned above, this dataset comes  with a set of binary attributes describing each image. Among them, we consider the blurriness as a non-sensitive information, so that the percentage of sharp images within the private dataset can be shared with the server. Therefore, we define $c_{3,a} = \xi(\mathcal{D}_a)/\sum_{i \in \mathcal{A}^-} \xi(\mathcal{D}_i)$, where $\xi$ measures the fraction of sharp images in a specific dataset. This criterion (\textbf{IS}) gives us an idea of the quality of the images in the dataset.

\subsection{Setting}

\paragraph{Implementation} 
We implemented our framework using Tensorflow~\cite{tensorflow2015-whitepaper} interface for Python. The operations and the communication between the participants of the federation (one server $S$ and $|\mathcal{A}|$ clients) was simulated by using a single machine\footnote{A public implementation of our framework is available at \url{https://github.com/sisinflab/ClientAware-FL}.}.

\paragraph{Hyperparameters}
We set the hyperparameters for the whole set of our experiments as follows, also guided by the results obtained in~\cite{mcmahan2017communication}. As for the FedAvg client fraction parameter, in each round of communication only 10\% of clients are selected to perform the computation, so that $|\mathcal{A}^-| = |\mathcal{A}|/10$. For what concerns the parameters of stochastic gradient decent (SGD), we used full batch approach and we set the number of local epochs equal to 5, i.e., each client takes 5 epochs of gradient descent during training. Moreover, we chose the learning rate based on a grid search by looking for the value which makes it possible to first reach the target accuracy in 50\% of devices with the baseline approach (dataset size as the only criterion). Finally, we set the maximum number of rounds of communication per each experiment to 1000 for MNIST and to 100 for CelebA.

\paragraph{Evaluation} The classification model we built have been evaluated with respect to classification accuracy, i.e.,

\begin{equation}
    \text{accuracy} = \frac{\text{\# correct predictions}}{\text{\# total predictions}}.
\end{equation}

This metric has been computed on each device over the private local test set. 
Based on LEAF framework~\cite{caldas2018leaf} --- which provides reproducible reference implementations and datasets for FL, as well as system and statistical metrics ---, we estimate a global accuracy by averaging local accuracy values and weighting them based on local test set size.

Moreover, we improve the validation of the FL setting by using an approach which offers an overview of the whole training performances, instead of metrics describing a single round of communication. More specifically, we measure the number of round of communication required to allow a certain percentage of devices, which participate to the federation process, to reach a target accuracy (e.g., 75\% or 80\%), since this measurement is able to fairly show how effective and efficient is the model across the devices.

\section{Results and Discussion}
\label{subsec:eval_FL_proposed}
In this section, we show and discuss the results of the experiments by considering both the evaluation approaches previously mentioned.

Tables \ref{tbl:iid}, \ref{tbl:noniid}, and \ref{tbl:celeba} show in the gray cells the number of rounds of communication required to the baseline (only DS) to allow certain fractions of devices to reach a specific percentage of the overall accuracy, for MNIST IID distributed, MNIST Non-IID distributed, and CelebA, respectively.
In detail, as target accuracies, we have considered 70\%, 80\%, 90\%, and 95\% of prediction accuracy for MNIST datasets, and 70\%, 80\%, and 85\% for CelebA dataset. 
The remaining rows of the tables show the gain in terms of rounds of communication of each experimental setting against the baseline.
Therefore, we compute each row as the difference between the results of the dataset size criterion model, and the results of the corresponding model. The higher the positive value, the better is the approach compared to the baseline.  Moreover, for each row and target accuracy, we show the average rounds of communication gained by considering all the fractions of devices. 

For each target accuracy, we show how many training rounds are needed to grant that accuracy for 10\% up to 90\% of devices. 

To have a comprehensive view of the effects of each criterion and their combinations, we consider the individual criteria and their combinations separately.
Specifically, we first show the results of the experiments realized by exploiting a single criterion (i.e., in Eq.~\ref{eq:pasi}).
Then, we show the results for the six priority permutations of the three criteria.

Moreover, we provide for each dataset the confusion matrices (Tables~\ref{tbl:conf-iid}, \ref{tbl:conf-noniid}, and \ref{tbl:conf-celeba}) which summarize the classification outcomes (correctly classified vs. misclassified) of each experiment against the baseline (only DS criterion). They have been obtained by comparing the best outcome of each experiment in order to highlight the significance of results irrespectively of the accuracy alignment presented in Tables \ref{tbl:iid}, \ref{tbl:noniid}, and \ref{tbl:celeba}.

Finally, we also show the results in terms of global test accuracy while considering the rounds of communication.
In this respect, we have measured global accuracy as the average of the local test accuracy values weighted by the local test size. 

\begin{figure}[ht]
\small
\captionsetup[subfigure]{aboveskip=2pt,belowskip=4pt}
\centering
\begin{tikzpicture}[trim axis left, trim axis right]
\footnotesize
\begin{axis}
[
    yticklabel style={
            /pgf/number format/fixed,
            /pgf/number format/precision=2
    },
    scaled y ticks=false,
    every axis x label/.style={at={(current axis.right of origin)},anchor=north west},
    ylabel near ticks,
    legend pos=south east,
    legend cell align={left},
]
\addplot[red, line width=0.6pt, mark size=1pt, restrict x to domain=0:100] table[x=num_round, y=0, col sep=semicolon]{celeba_aggregated.csv};
\addlegendentry{DS};
\addplot[green, line width=0.6pt, restrict x to domain=0:100] table[x=num_round, y=1, col sep=semicolon]{celeba_aggregated.csv};
\addlegendentry{IS};
\addplot[blue, line width=0.6pt, restrict x to domain=0:100] table[x=num_round, y=2, col sep=semicolon]{celeba_aggregated.csv};
\addlegendentry{CB};
\addplot[SpringGreen, line width=0.6pt ,restrict x to domain=0:100] table[x=num_round, y=3, col sep=semicolon]{celeba_aggregated.csv};
\addlegendentry{DS $\succ$ IS $\succ$ CB}
\addplot[Aquamarine, line width=0.6pt, restrict x to domain=0:100] table[x=num_round, y=4, col sep=semicolon]{celeba_aggregated.csv};
\addlegendentry{DS $\succ$ CB $\succ$ IS}
\addplot[Tan, line width=0.6pt, restrict x to domain=0:100] table[x=num_round, y=5, col sep=semicolon]{celeba_aggregated.csv};
\addlegendentry{IS $\succ$ DS $\succ$ CB}
\addplot[Violet, line width=0.6pt, restrict x to domain=0:100] table[x=num_round, y=6, col sep=semicolon]{celeba_aggregated.csv};
\addlegendentry{CB $\succ$ DS $\succ$ IS}
\addplot[VioletRed, line width=0.6pt, restrict x to domain=0:100] table[x=num_round, y=7, col sep=semicolon]{celeba_aggregated.csv};
\addlegendentry{IS $\succ$ CB $\succ$ DS}
\addplot[Yellow, line width=0.6pt, restrict x to domain=0:100] table[x=num_round, y=8, col sep=semicolon]{celeba_aggregated.csv};
\addlegendentry{CB $\succ$ IS $\succ$ DS}

\end{axis}

\end{tikzpicture}
\caption{Results for CelebA dataset. The plot provides the measure of test global accuracy during training for each round of communication (on horizontal axis).}
\label{fig:celeba}
\end{figure}
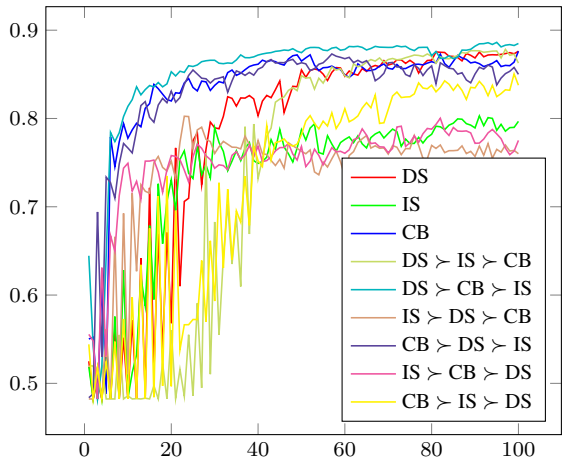

In detail, Figures \ref{fig:iid}, \ref{fig:noniid}, and \ref{fig:celeba}, show these curves for MNIST IID distributed, MNIST Non-IID distributed, and CelebA, respectively.

{
\captionsetup[table]{labelformat=empty}
\begin{table*}[htbp]
    \caption*{}
\end{table*}
}

\begin{table*}[htbp]
\caption{Results for MNIST dataset with IID distribution for each experiment (the symbol $\succ$ denotes the priority relation). 
The row DS (baselines) provides the number of rounds of communication needed to make the percentage of devices specified in the columns reach the specified target accuracy. The remaining rows show the gain in the number of communication rounds with respect to the baseline (the greater the better). The column Avg. shows the average gain for the experiment with all the percentages of devices.}
\begin{center}
\setlength\tabcolsep{1.9pt}
\begin{tabular}{|l|cccccccccc|cccccccccc|}
\hline

  & \multicolumn{ 10}{c|}{\textbf{Target accuracy 70\%}} & \multicolumn{ 10}{c|}{\textbf{Target accuracy 80\%}} \\ \hline
\multicolumn{1}{|l|}{\textbf{Experiment}} & \textbf{10\%} & \textbf{20\%} & \textbf{30\%} & \textbf{40\%} & \textbf{50\%} & \textbf{60\%} & \textbf{70\%} & \textbf{80\%} & \textbf{90\%} & \textbf{Avg.} & \textbf{10\%} & \textbf{20\%} & \textbf{30\%} & \textbf{40\%} & \textbf{50\%} & \textbf{60\%} & \textbf{70\%} & \textbf{80\%} & \textbf{90\%} & \textbf{Avg.} \\ \hline
\rowcolor{Gray!30!} \textbf{DS (baseline)} & 1 & 1 & 1 & 1 & 1 & 4 & 4 & 4 & 5 &  & 1 & 2 & 5 & 5 & 5 & 5 & 5 & 5 & 5 &  \\ \hline
\textbf{LD} & 0 & 0 & 0 & -2 & -2 & 1 & 1 & 1 & 0 & \multicolumn{1}{r|}{-0,11} & -2 & -1 & 0 & 0 & 0 & 0 & 0 & 0 & 0 & \multicolumn{1}{l|}{-0,33} \\ \hline
\textbf{MW} & 0 & 0 & 0 & 0 & 0 & 3 & 0 & 0 & 0 & \multicolumn{1}{r|}{0,33} & 0 & -2 & 1 & 0 & 0 & 0 & 0 & 0 & 0 & \multicolumn{1}{l|}{-0,11} \\ \hline
\textbf{DS $\succ$ LD $\succ$ MW} & 0 & 0 & -1 & -3 & -3 & 0 & 0 & 0 & 0 & \multicolumn{1}{r|}{-0,78} & -3 & -2 & 0 & 0 & 0 & 0 & 0 & 0 & 0 & \multicolumn{1}{l|}{-0,56} \\ \hline
\textbf{DS $\succ$ MW $\succ$ LD} & 0 & 0 & 0 & 0 & 0 & 0 & 0 & 0 & 1 & \multicolumn{1}{r|}{0,11} & -3 & -2 & 1 & 1 & 1 & 1 & 1 & 1 & 1 & \multicolumn{1}{l|}{0,22} \\ \hline
\textbf{LD $\succ$ DS $\succ$ MW} & 0 & 0 & 0 & 0 & 0 & 2 & 0 & 0 & 1 & \multicolumn{1}{r|}{0,33} & -1 & -2 & 1 & 1 & 1 & 0 & 0 & 0 & -1 & \multicolumn{1}{l|}{-0,11} \\ \hline
\textbf{MW $\succ$ DS $\succ$ LD} & 0 & 0 & 0 & -3 & -3 & 0 & 0 & 0 & 1 & \multicolumn{1}{r|}{-0,56} & -3 & -2 & 1 & 1 & 1 & 1 & 0 & 0 & 0 & \multicolumn{1}{l|}{-0,11} \\ \hline
\textbf{LD $\succ$ MW $\succ$ DS} & 0 & 0 & 0 & 0 & 0 & 3 & 3 & 3 & 1 & \multicolumn{1}{r|}{1,11} & 0 & 1 & 4 & 1 & 1 & 1 & 1 & 1 & 1 & \multicolumn{1}{l|}{1,22} \\ \hline
\textbf{MW $\succ$ LD $\succ$ DS} & 0 & 0 & 0 & 0 & 0 & 1 & 1 & 1 & 1 & \multicolumn{1}{r|}{0,44} & -2 & -1 & 1 & 1 & 0 & 0 & 0 & 0 & 0 & \multicolumn{1}{l|}{-0,11} \\ \hline
\end{tabular}

\medskip

\begin{tabular}{|l|cccccccccc|cccccccccc|}
\hline
 & \multicolumn{ 10}{c|}{\textbf{Target accuracy 90\%}} & \multicolumn{ 10}{c|}{\textbf{Target accuracy 95\%}} \\ \hline
\multicolumn{1}{|l|}{\textbf{Experiment}} & \textbf{10\%} & \textbf{20\%} & \textbf{30\%} & \textbf{40\%} & \textbf{50\%} & \textbf{60\%} & \textbf{70\%} & \textbf{80\%} & \textbf{90\%} & \textbf{Avg.} & \textbf{10\%} & \textbf{20\%} & \textbf{30\%} & \textbf{40\%} & \textbf{50\%} & \textbf{60\%} & \textbf{70\%} & \textbf{80\%} & \textbf{90\%} & \textbf{Avg.} \\ \hline
\rowcolor{Gray!30!} \textbf{DS (baseline)} & 5 & 5 & 5 & 5 & 7 & 7 & 10 & 11 & 14 &  & 5 & 7 & 10 & 11 & 16 & 21 & 24 & 38 & 56 &  \\ \hline
\textbf{LD} & 0 & 0 & 0 & 0 & 0 & 0 & 1 & -1 & -1 & \multicolumn{1}{r|}{-0,11} & 0 & 0 & 2 & 0 & -2 & 1 & -3 & 0 & -3 & \multicolumn{1}{r|}{-0,56} \\ \hline
\textbf{MW} & 0 & 0 & 0 & 0 & 0 & 0 & 0 & -1 & -1 & \multicolumn{1}{r|}{-0,22} & 0 & 0 & 2 & -4 & 1 & -1 & -4 & 1 & -3 & \multicolumn{1}{r|}{-0,89} \\ \hline
\textbf{DS $\succ$ LD $\succ$ MW} & 0 & 0 & 0 & 0 & 2 & 0 & -1 & -1 & 0 & \multicolumn{1}{r|}{0,00} & 0 & 0 & 2 & 0 & -3 & 0 & -5 & 4 & -10 & \multicolumn{1}{r|}{-1,33} \\ \hline
\textbf{DS $\succ$ MW $\succ$ LD} & 1 & 1 & 1 & -1 & 0 & 0 & 1 & -2 & -1 & \multicolumn{1}{r|}{0,00} & 1 & 0 & 1 & -1 & -1 & 0 & -1 & 3 & -14 & \multicolumn{1}{r|}{-1,33} \\ \hline
\textbf{LD $\succ$ DS $\succ$ MW} & 1 & 0 & 0 & -3 & -1 & -1 & 1 & 0 & -2 & \multicolumn{1}{r|}{-0,56} & 0 & -1 & 2 & -2 & -1 & -2 & -2 & -2 & -3 & \multicolumn{1}{r|}{-1,22} \\ \hline
\textbf{MW $\succ$ DS $\succ$ LD} & 0 & 0 & -1 & -1 & 0 & 0 & 2 & -1 & 1 & \multicolumn{1}{r|}{0,00} & -1 & 0 & 2 & -1 & 2 & 2 & 1 & 4 & -4 & \multicolumn{1}{r|}{0,56} \\ \hline
\textbf{LD $\succ$ MW $\succ$ DS} & 1 & 1 & 1 & 1 & 2 & -1 & 2 & 1 & 2 & \multicolumn{1}{r|}{1,11} & 1 & 1 & 2 & 2 & -1 & 2 & -1 & 0 & -26 & \multicolumn{1}{r|}{-2,22} \\ \hline
\textbf{MW $\succ$ LD $\succ$ DS} & 0 & 0 & 0 & 0 & 2 & -1 & 1 & 0 & -1 & \multicolumn{1}{r|}{0,11} & 0 & 0 & 2 & -1 & -1 & -1 & -3 & 4 & -16 & \multicolumn{1}{r|}{-1,78} \\ \hline
\end{tabular}

\end{center}
\label{tbl:iid}
\end{table*}

\begin{table*}[htbp]
\setlength\tabcolsep{3.4pt}
\caption{Confusion Matrices for MNIST dataset with IID distribution for each experiment. 
For each matrix, the rows refer to the number of samples that are misclassified (\xmark), or correctly classified (\cmark).
The columns denote the criteria that are compared with the baseline. The symbol $\succ$ denotes the priority relation.}
\begin{tabular}{|l|cc|cc|cc|cc|cc|cc|cc|cc|}
\hline
 & \multicolumn{2}{c|}{\textbf{LD}} &  \multicolumn{2}{c|}{\textbf{MW}} & \multicolumn{2}{c|}{\textbf{DS$\succ$LD$\succ$MW}} &  \multicolumn{2}{c|}{\textbf{DS$\succ$MW$\succ$LD}} & \multicolumn{2}{c|}{\textbf{LD$\succ$DS$\succ$MW}} & \multicolumn{2}{c|}{\textbf{MW$\succ$DS$\succ$LD}}  & \multicolumn{2}{c|}{\textbf{LD$\succ$MW$\succ$DS}} & \multicolumn{2}{c|}{\textbf{MW$\succ$LD$\succ$DS}}  \\ \hline
& \xmark & \cmark & \xmark & \cmark & \xmark & \cmark & \xmark & \cmark & \xmark & \cmark & \xmark & \cmark & \xmark & \cmark & \xmark & \cmark   \\ \hline
\textbf{DS}\hspace{1em}\xmark & 0 & 98 & 0 & 98 & 2 & 96 & 0 & 98 & 2 & 96 & 0 & 98 & 4 & 94 & 0 & 98 \\ \hline
\textbf{DS}\hspace{1em}\cmark & 106 & 9796 & 107 & 9795 & 106 & 9796 & 128 & 9774 & 118 & 9784 & 140 & 9762 & 107 & 9795 & 122 & 9780 \\ \hline
\end{tabular}
\label{tbl:conf-iid}
\end{table*}

\begin{table*}[htbp]
\caption{Results for MNIST dataset with Non-IID distribution for each experiment (the symbol $\succ$ denotes the priority relation).
The row DS (baselines) provides the number of rounds of communication needed to make the percentage of devices specified in the columns reach the specified target accuracy. The remaining rows show the gain in the number of communication rounds with respect to the baseline  (the greater the better). The column Avg. shows the average gain for the experiment with all the percentages of devices.}
\begin{center}
\setlength\tabcolsep{1.9pt}

\begin{tabular}{|l|cccccccccc|cccccccccc|}
\hline
 & \multicolumn{ 10}{c|}{\textbf{Target accuracy 70\%}} & \multicolumn{ 10}{c|}{\textbf{Target accuracy 80\%}} \\ \hline
\multicolumn{1}{|l|}{\textbf{Experiment}} & \textbf{10\%} & \textbf{20\%} & \textbf{30\%} & \textbf{40\%} & \textbf{50\%} & \textbf{60\%} & \textbf{70\%} & \textbf{80\%} & \textbf{90\%} & \textbf{Avg.} & \textbf{10\%} & \textbf{20\%} & \textbf{30\%} & \textbf{40\%} & \textbf{50\%} & \textbf{60\%} & \textbf{70\%} & \textbf{80\%} & \textbf{90\%} & \textbf{Avg.} \\ \hline
\rowcolor{Gray!30!} \textbf{DS (baseline)} & 9 & 9 & 9 & 10 & 10 & 10 & 10 & 10 & 10 &  & 10 & 10 & 10 & 10 & 10 & 10 & 16 & 20 & 20 &  \\ \hline
\textbf{LD} & 3 & 1 & 1 & 2 & 2 & 2 & 2 & 2 & -1 & \multicolumn{1}{r|}{1,56} & 2 & 2 & 2 & -1 & -3 & -3 & 3 & 7 & 2 & \multicolumn{1}{r|}{1,22} \\ \hline
\textbf{MW} & 4 & 4 & 4 & 5 & 0 & 0 & 0 & 0 & 0 & \multicolumn{1}{r|}{1,89} & 0 & 0 & 0 & 0 & -6 & -6 & 0 & 1 & 1 & \multicolumn{1}{r|}{-1,11} \\ \hline
\textbf{DS $\succ$ LD $\succ$ MW} & 2 & 2 & 2 & 2 & 2 & 2 & 2 & 1 & -2 & \multicolumn{1}{r|}{1,44} & 2 & 1 & 1 & 1 & -3 & -3 & 1 & 5 & 5 & \multicolumn{1}{r|}{1,11} \\ \hline
\textbf{DS $\succ$ MW $\succ$ LD} & 0 & 0 & 0 & 0 & 0 & 0 & 0 & 0 & -4 & \multicolumn{1}{r|}{-0,44} & 1 & 0 & 0 & -4 & -7 & -7 & -1 & 3 & 2 & \multicolumn{1}{r|}{-1,44} \\ \hline
\textbf{LD $\succ$ DS $\succ$ MW} & 2 & 2 & 1 & 2 & 2 & 2 & 2 & 2 & 2 & \multicolumn{1}{r|}{1,89} & 2 & 2 & 2 & -1 & -1 & -1 & 5 & 6 & 5 & \multicolumn{1}{r|}{2,11} \\ \hline
\textbf{MW $\succ$ DS $\succ$ LD} & 1 & 0 & 0 & 0 & 0 & 0 & 0 & -1 & -1 & \multicolumn{1}{r|}{-0,11} & 0 & 0 & -1 & -1 & -1 & -1 & 2 & 2 & 2 & \multicolumn{1}{r|}{0,22} \\ \hline
\textbf{LD $\succ$ MW $\succ$ DS} & 0 & 0 & 0 & 1 & 1 & -2 & -2 & -2 & -2 & \multicolumn{1}{r|}{-0,67} & 1 & -2 & -2 & -2 & -3 & -5 & 1 & 0 & 0 & \multicolumn{1}{r|}{-1,33} \\ \hline
\textbf{MW $\succ$ LD $\succ$ DS} & 1 & 1 & 1 & 2 & 1 & 1 & 1 & -3 & -3 & \multicolumn{1}{r|}{0,22} & 1 & 1 & -3 & -3 & -3 & -5 & 0 & 4 & 3 & \multicolumn{1}{r|}{-0,56} \\ \hline
\end{tabular}

\medskip

{
\setlength\tabcolsep{1.8pt}
\begin{tabular}{|l|cccccccccc|cccccccccc|}
\hline
 & \multicolumn{ 10}{c|}{\textbf{Target accuracy 90\%}} & \multicolumn{ 10}{c|}{\textbf{Target accuracy 95\%}} \\ \hline
\multicolumn{1}{|l|}{\textbf{Experiment}} & \textbf{10\%} & \textbf{20\%} & \textbf{30\%} & \textbf{40\%} & \textbf{50\%} & \textbf{60\%} & \textbf{70\%} & \textbf{80\%} & \textbf{90\%} & \textbf{Avg.} & \textbf{10\%} & \textbf{20\%} & \textbf{30\%} & \textbf{40\%} & \textbf{50\%} & \textbf{60\%} & \textbf{70\%} & \textbf{80\%} & \textbf{90\%} & \textbf{Avg.} \\ \hline
\rowcolor{Gray!30!} \textbf{DS (baseline)} & 10 & 16 & 20 & 20 & 23 & 33 & 35 & 44 & 69 &  & 20 & 23 & 34 & 49 & 66 & 84 & 106 & 137 & 222 &  \\ \hline
\textbf{LD} & -1 & 3 & 2 & 2 & -2 & 2 & -4 & -5 & 8 & \multicolumn{1}{r|}{0,56} & 2 & -2 & -7 & -3 & 3 & -5 & -1 & -11 & -23 & \multicolumn{1}{r|}{-5,22} \\ \hline
\textbf{MW} & -9 & -3 & 1 & -7 & -15 & -8 & -11 & -13 & -2 & \multicolumn{1}{r|}{-7,44} & 1 & -15 & -7 & -13 & -10 & -15 & -23 & -11 & -14 & \multicolumn{1}{r|}{-11,89} \\ \hline
\textbf{DS $\succ$ LD $\succ$ MW} & -3 & 1 & 5 & 3 & 4 & 5 & -4 & 1 & 4 & \multicolumn{1}{r|}{1,78} & 5 & -3 & -5 & 0 & -11 & -13 & -5 & -19 & -11 & \multicolumn{1}{r|}{-6,89} \\ \hline
\textbf{DS $\succ$ MW $\succ$ LD} & -7 & -1 & 2 & -5 & -7 & -6 & -9 & -12 & 9 & \multicolumn{1}{r|}{-4,00} & 2 & -7 & -5 & -5 & -14 & -14 & -6 & -16 & 12 & \multicolumn{1}{r|}{-5,89} \\ \hline
\textbf{LD $\succ$ DS $\succ$ MW} & 2 & 5 & 5 & -1 & -6 & 0 & -3 & -4 & 1 & \multicolumn{1}{r|}{-0,11} & 5 & -6 & 1 & -2 & -4 & 1 & 12 & -2 & 3 & \multicolumn{1}{r|}{0,89} \\ \hline
\textbf{MW $\succ$ DS $\succ$ LD} & -1 & 5 & 2 & -3 & -2 & 3 & -4 & -4 & 9 & \multicolumn{1}{r|}{0,56} & 2 & 0 & 3 & 0 & 4 & 0 & 2 & 10 & -6 & \multicolumn{1}{r|}{1,67} \\ \hline
\textbf{LD $\succ$ MW $\succ$ DS} & -3 & -4 & 0 & -4 & -3 & 1 & -1 & -4 & -1 & \multicolumn{1}{r|}{-2,11} & 0 & -1 & -1 & -10 & -3 & -2 & -5 & 5 & -6 & \multicolumn{1}{r|}{-2,56} \\ \hline
\textbf{MW $\succ$ LD $\succ$ DS} & -3 & 0 & 3 & 1 & -1 & 4 & 3 & 7 & 14 & \multicolumn{1}{r|}{3,11} & 4 & 3 & 2 & 1 & 0 & 9 & -9 & -5 & 37 & \multicolumn{1}{r|}{4,67} \\ \hline
\end{tabular}
}
\end{center}
\label{tbl:noniid}
\end{table*}

\begin{table*}[htbp]
\setlength\tabcolsep{3.4pt}
\caption{Confusion Matrices for MNIST dataset with Non-IID distribution for each experiment. 
For each matrix, the rows refer to the number of samples that are misclassified (\xmark), or correctly classified (\cmark).
The columns denote the criteria that are compared with the baseline. The symbol $\succ$ denotes the priority relation.}
\begin{tabular}{|l|cc|cc|cc|cc|cc|cc|cc|cc|}
\hline
 & \multicolumn{2}{c|}{\textbf{LD}} &  \multicolumn{2}{c|}{\textbf{MW}} & \multicolumn{2}{c|}{\textbf{DS$\succ$LD$\succ$MW}} &  \multicolumn{2}{c|}{\textbf{DS$\succ$MW$\succ$LD}} & \multicolumn{2}{c|}{\textbf{LD$\succ$DS$\succ$MW}} & \multicolumn{2}{c|}{\textbf{MW$\succ$DS$\succ$LD}}  & \multicolumn{2}{c|}{\textbf{LD$\succ$MW$\succ$DS}} & \multicolumn{2}{c|}{\textbf{MW$\succ$LD$\succ$DS}}  \\ \hline
& \xmark & \cmark & \xmark & \cmark & \xmark & \cmark & \xmark & \cmark & \xmark & \cmark & \xmark & \cmark & \xmark & \cmark & \xmark & \cmark  \\ \hline
\textbf{DS}\hspace{1em}\xmark & 1 & 130 & 2 & 129 & 1 & 130 & 2 & 129 & 1 & 130 & 0 & 131 & \multicolumn{1}{r}{2} & 129 & 0 & 131 \\ \hline
\textbf{DS}\hspace{1em}\cmark & 146 & 9723 & 142 & 9727 & 119 & 9750 & 127 & 9742 & 103 & 9766 & 104 & 9765 & 124 & 9745 & 119 & 9750 \\ \hline
\end{tabular}
\label{tbl:conf-noniid}
\end{table*}

\begin{table*}[htbp]
\caption{Results for CelebA dataset for each experiment (the symbol $\succ$ denotes the priority relation). The row DS (baselines) provides the number of rounds of communication needed to make the percentage of devices specified in the columns reach the specified target accuracy. The remaining rows show the gain in the number of communication rounds with respect to the baseline  (the greater the better). The column Avg. shows the average gain for the experiment with all the percentages of devices.
Runs that did not reach the target accuracy for the specified percentage of devices in the 100 allowed rounds
have been obtained considering such limit value.}
\begin{center}
\setlength\tabcolsep{0.67pt}

\begin{tabular}{|l|ccccccccc|cccccccc|cccccc|}
\hline
 & \multicolumn{ 9}{c|}{\textbf{Target accuracy 70\%}} & \multicolumn{ 8}{c|}{\textbf{Target accuracy 80\%}} & \multicolumn{ 6}{c|}{\textbf{Target accuracy 85\%}} \\ \hline
\textbf{Experiment} & \textbf{10\%} & \textbf{20\%} & \textbf{30\%} & \textbf{40\%} & \textbf{50\%} & \textbf{60\%} & \textbf{70\%} & \textbf{80\%} & \textbf{Avg.} & \textbf{10\%} & \textbf{20\%} & \textbf{30\%} & \textbf{40\%} & \textbf{50\%} & \textbf{60\%} & \textbf{70\%} & \multicolumn{1}{l|}{\textbf{Avg.}} & \textbf{10\%} & \textbf{20\%} & \textbf{30\%} & \textbf{40\%} & \textbf{50\%} & \multicolumn{1}{l|}{\textbf{Avg.}} \\ \hline
\rowcolor{Gray!30!} \textbf{DS (baseline)} & 1 & 1 & 1 & 13 & 15 & 21 & 32 & 58 &  & 1 & 1 & 15 & 21 & 26 & 44 & --- &  & 1 & 15 & 21 & 29 & 50 &  \\ \hline
\textbf{IS} & 0 & 0 & -2 & 4 & -2 & -1 & -59 & -42 & \multicolumn{1}{r|}{-11,33} & 0 & -2 & 0 & 4 & -4 & -56 & --- & \multicolumn{1}{r|}{-6,44} & -1 & 2 & 4 & -1 & -50 & \multicolumn{1}{r|}{-5,11} \\ \hline
\textbf{CB} & 0 & 0 & 0 & 7 & 9 & 15 & 18 & 20 & \multicolumn{1}{r|}{7,67} & 0 & 0 & 9 & 15 & 17 & 28 & --- & \multicolumn{1}{r|}{7,67} & 0 & 9 & 15 & 15 & 26 & \multicolumn{1}{r|}{7,22} \\ \hline
\textbf{DS $\succ$ IS $\succ$ CB} & 0 & 0 & 0 & -15 & -13 & -7 & -12 & 1 & \multicolumn{1}{r|}{-5,11} & 0 & 0 & -13 & -7 & -11 & -3 & --- & \multicolumn{1}{r|}{-3,78} & 0 & -13 & -7 & -15 & 0 & \multicolumn{1}{r|}{-3,89} \\ \hline
\textbf{DS $\succ$ CB $\succ$ IS} & 0 & 0 & 0 & 12 & 9 & 15 & 22 & 34 & \multicolumn{1}{r|}{10,22} & 0 & 0 & 10 & 15 & 18 & 32 & 43 & \multicolumn{1}{r|}{13,11} & 0 & 10 & 15 & 20 & 30 & \multicolumn{1}{r|}{8,33} \\ \hline
\textbf{IS $\succ$ DS $\succ$ CB} & 0 & 0 & -2 & 6 & 6 & 5 & 9 & -42 & \multicolumn{1}{r|}{-2,00} & 0 & -2 & 6 & 10 & 3 & -56 & --- & \multicolumn{1}{r|}{-4,33} & 0 & 6 & 8 & 6 & -50 & \multicolumn{1}{r|}{-3,33} \\ \hline
\textbf{CB $\succ$ DS $\succ$ IS} & 0 & 0 & -2 & 10 & 6 & 13 & 19 & 18 & \multicolumn{1}{r|}{7,11} & 0 & -2 & 12 & 16 & 14 & 21 & --- & \multicolumn{1}{r|}{6,78} & 0 & 12 & 16 & 16 & 26 & \multicolumn{1}{r|}{7,78} \\ \hline
\textbf{IS $\succ$ CB $\succ$ DS} & 0 & 0 & 0 & 9 & 6 & 12 & -68 & -42 & \multicolumn{1}{r|}{-9,22} & 0 & -3 & 9 & 12 & -21 & -56 & --- & \multicolumn{1}{r|}{-6,56} & 0 & 9 & 12 & -35 & -50 & \multicolumn{1}{r|}{-7,11} \\ \hline
\textbf{CB $\succ$ IS $\succ$ DS} & 0 & 0 & -2 & 0 & 6 & -18 & -20 & -42 & \multicolumn{1}{r|}{-8,44} & 0 & -2 & 0 & 2 & -16 & -28 & --- & \multicolumn{1}{r|}{-4,89} & 0 & 0 & -10 & -13 & -33 & \multicolumn{1}{r|}{-6,22} \\ \hline
\end{tabular}
\end{center}
\label{tbl:celeba}
\end{table*}

\begin{table*}[htbp]
\setlength\tabcolsep{4.2pt}
\caption{Confusion Matrices for CelebA dataset for each experiment. 
For each matrix, the rows refer to the number of samples that are misclassified (\xmark), or correctly classified (\cmark).
The columns denote the criteria that are compared with the baseline. The symbol $\succ$ denotes the priority relation.}
\begin{tabular}{|l|cc|cc|cc|cc|cc|cc|cc|cc|}
\hline
 & \multicolumn{2}{c|}{\textbf{IS}} &  \multicolumn{2}{c|}{\textbf{CB}} & \multicolumn{2}{c|}{\textbf{DS$\succ$IS$\succ$CB}} &  \multicolumn{2}{c|}{\textbf{DS$\succ$CB$\succ$IS}} & \multicolumn{2}{c|}{\textbf{IS$\succ$DS$\succ$CB}} & \multicolumn{2}{c|}{\textbf{CB$\succ$DS$\succ$IS}}  & \multicolumn{2}{c|}{\textbf{IS$\succ$CB$\succ$DS}} & \multicolumn{2}{c|}{\textbf{CB$\succ$IS$\succ$DS}}  \\ \hline
& \xmark & \cmark & \xmark & \cmark & \xmark & \cmark & \xmark & \cmark & \xmark & \cmark & \xmark & \cmark & \xmark & \cmark & \xmark & \cmark  \\ \hline
\textbf{DS}\hspace{1em}\xmark & 312 & 2081 & 610 & 1783 & 275 & 2118 & 296 & 2097 & 331 & 2062 & 618 & 1775 & 416 & 1977 & 612 & 1781 \\ \hline
\textbf{DS}\hspace{1em}\cmark & 2080 & 14846 & 3197 & 13729 & 1923 & 15003 & 2056 & 14870 & 2121 & 14805 & 3197 & 13729 & 2473 & 14453 & 3247 & 13679 \\ \hline
\end{tabular}
\label{tbl:conf-celeba}
\end{table*}

\subsection{Effects of individual criteria}
In the following, we discuss the experiments by investigating the effects of the individual criteria.
\paragraph{MNIST with IID distribution.}
Table~\ref{tbl:iid} and Figure~\ref{fig:iid} show the experimental results for MNIST considering IID distribution. 
The adoption of criteria that are different from the dataset size seems to have no effects, but some insignificant fluctuations. 
We expected this behavior since the dataset does not show statistical issues.
In detail, the local datasets have been created by randomly picking samples from the original dataset. 
Consequently, we may represent the divergence of the local models from the original model as a zero-mean Gaussian noise.
The same assumption holds for the label diversity.
Furthermore, by the design of the dataset, DS is equal for every client, and we also expect that, on average, LD is the same for all clients. 
As a consequence, MW can not play a significant role, since all clients train the model with the same distribution of data, on average.

In detail, by looking at the differences between the two criteria LD and MW against the baseline DS, we may notice that the gain in terms of rounds of communication for 70\%, 80\%, and 90\% target accuracy, considering all the percentages of clients, has an average very close to 0.
Conversely, both LD and MW show slightly worse results than DS for a target accuracy of 95\%, notably when we want to grant such accuracy to a very high number of devices (i.e., the last three columns of Table~\ref{tbl:iid}).

Lastly, the experiment shows that the effectiveness of criteria depends on the data distribution. 
The dataset shows how datasets built by randomly picking data from the original distribution do not generally need further statistical adjustments during the training phase. 
Therefore, we do not need to consider, in our approach, any particular statistical-based criteria. 
On the other hand, it is worth mentioning that the scenario is completely unrealistic in a federated scenario.
Indeed, since each user privately generates data, they will not show the same distribution, and they will be significantly different in terms of statistical properties.

\paragraph{MNIST with Non-IID distribution.}
Regarding DS, LD, and MW, Table \ref{tbl:noniid} shows that DS is the best criterion to reach the best overall system performance in a lower number of rounds.
In detail, DS seems to be in trouble in the first rounds to achieve an acceptable degree of accuracy. In fact, if we look at 70\% and 80\% of the accuracy, LD and MW reach those goals faster. In this sense, we may notice an average of 1-2 rounds gained by LD and MW.
However, if we need a higher degree of accuracy (90\% and 95\%), DS generally reaches that accuracy from 5 to 12 rounds before, on average.
In this respect, we may notice that, without considering the last three columns of 95\%, DS and LD show approximately the same performance, with an advantage of 0.5 rounds on average for LD. On the other hand, in the same scenario, MW is 3.6 rounds slower on average.
Nonetheless, the remaining three last columns show an average advantage of DS of 12 and 16 rounds for LD and MW, respectively.

\paragraph{CelebA.}
Results in Table \ref{tbl:celeba} show that the three criteria behave in a significantly different way.
In this experiment, we note that CB performs much better than DS.
To get an impression, we may notice CB reaches 85\% of the overall accuracy for the 60\% of the devices in less than half of the rounds of DS.
Moreover, if we observe the CB row in Table \ref{tbl:celeba}, we may note that it reaches the same accuracy goals of DS much faster. In detail, CB reaches those goals 7.5 rounds before DS.
On the other side, IS is generally slower than DS. Indeed, it generally needs 7.6 rounds more than DS.
Another interesting insight is that the advantage of CB over DS progressively increases. In detail, CB shows an advantage of 8.6 rounds for the 70\% of accuracy, 11.5 rounds for 80\%, and 13 rounds for an accuracy of 85\%.

\begin{figure*}[ht]
\small
\begin{subfigure}{\columnwidth}
\centering
\begin{tikzpicture}[trim axis left]
\footnotesize
\begin{axis}
[
    yticklabel style={
            /pgf/number format/fixed,
            /pgf/number format/precision=2
    },
    scaled y ticks=false,
    every axis x label/.style={at={(current axis.right of origin)},anchor=north west},
    ylabel near ticks,
    legend pos=south east,
    legend cell align={left}
]
\addplot[red, line width=0.6pt, mark size=1pt, restrict x to domain=0:100] table[x=num_round, y=0, col sep=semicolon]{iid_aggregated.csv};
\addlegendentry{DS};
\addplot[green, line width=0.6pt, restrict x to domain=0:100] table[x=num_round, y=1, col sep=semicolon]{iid_aggregated.csv};
\addlegendentry{LD};
\addplot[blue, line width=0.6pt, restrict x to domain=0:100] table[x=num_round, y=2, col sep=semicolon]{iid_aggregated.csv};
\addlegendentry{MW};
\addplot[SpringGreen, line width=0.6pt ,restrict x to domain=0:100] table[x=num_round, y=3, col sep=semicolon]{iid_aggregated.csv};
\addlegendentry{DS $\succ$ LD $\succ$ MW};
\addplot[Aquamarine, line width=0.6pt, restrict x to domain=0:100] table[x=num_round, y=4, col sep=semicolon]{iid_aggregated.csv};
\addlegendentry{DS $\succ$ MW $\succ$ LD};
\addplot[Tan, line width=0.6pt, restrict x to domain=0:100] table[x=num_round, y=5, col sep=semicolon]{iid_aggregated.csv};
\addlegendentry{LD $\succ$ DS $\succ$ MW};
\addplot[Violet, line width=0.6pt, restrict x to domain=0:100] table[x=num_round, y=6, col sep=semicolon]{iid_aggregated.csv};
\addlegendentry{MW $\succ$ DS $\succ$ LD};
\addplot[VioletRed, line width=0.6pt, restrict x to domain=0:100] table[x=num_round, y=7, col sep=semicolon]{iid_aggregated.csv};
\addlegendentry{LD $\succ$ MW $\succ$ DS};
\addplot[Yellow, line width=0.6pt, restrict x to domain=0:100] table[x=num_round, y=8, col sep=semicolon]{iid_aggregated.csv};
\addlegendentry{MW $\succ$ LD $\succ$ LD};

\end{axis}
\end{tikzpicture}
\caption{IID distribution}
\label{fig:iid}
\end{subfigure}\hfill
\begin{subfigure}{\columnwidth}
\centering
\begin{tikzpicture}[trim axis left]
\footnotesize
\begin{axis}
[
    yticklabel style={
            /pgf/number format/fixed,
            /pgf/number format/precision=2
    },
    scaled y ticks=false,
    every axis x label/.style={at={(current axis.right of origin)},anchor=north west},
    ylabel near ticks,
    legend pos=south east,
    legend cell align={left},
]
\addplot[red, line width=0.6pt, mark size=1pt, restrict x to domain=0:100] table[x=num_round, y=0, col sep=semicolon]{non-iid_aggregated.csv};
\addlegendentry{DS};
\addplot[Tan, line width=0.6pt, restrict x to domain=0:100] table[x=num_round, y=5, col sep=semicolon]{non-iid_aggregated.csv};
\addlegendentry{LD $\succ$ DS $\succ$ MW};
\addplot[Violet, line width=0.6pt, restrict x to domain=0:100] table[x=num_round, y=6, col sep=semicolon]{non-iid_aggregated.csv};
\addlegendentry{MW $\succ$ DS $\succ$ LD};
\addplot[Yellow, line width=0.6pt, restrict x to domain=0:100] table[x=num_round, y=8, col sep=semicolon]{non-iid_aggregated.csv};
\addlegendentry{MW $\succ$ LD $\succ$ LD};

\end{axis}
\end{tikzpicture}
\caption{Non-IID distribution}
\label{fig:noniid}
\end{subfigure}
\caption{Results for MNIST dataset. The plot provides the measure of test global accuracy during training for each round of communication (on horizontal axis).}
\end{figure*}
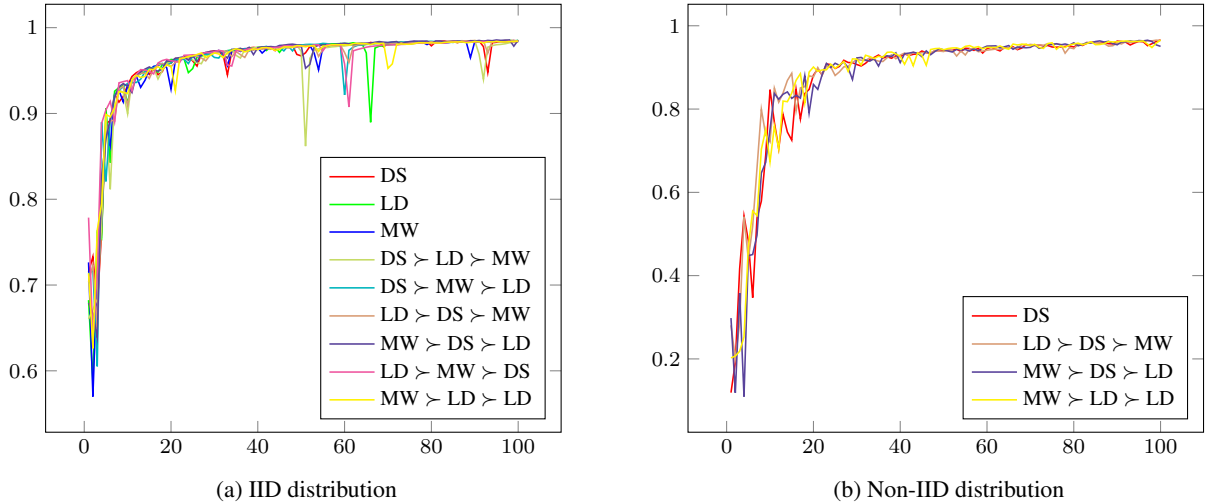

\subsection{Effects of combined criteria}
In this section, we focus on the effects of the combination of the different criteria. 
Eventually, in Section \ref{sec:discussion}, we discuss the different findings of the experiments, and we draw some general remarks.
\paragraph{MNIST with IID distribution.}
In the lower section, Table \ref{tbl:iid} shows the effects of the six priority permutations of the criteria. 
We may suppose that, since the individual criteria did not give any boost to the training, also their combination should have no effect. 
As an example, in some situations --- and this is very clear in LD $\succ$ MW $\succ$ DS ---  we can observe a boost for lower target value of accuracy (70\%, 80\%, and 90\%) and a slowdown in reaching the target accuracy of 95\% for a high number of devices. 
Analogously to the individual criteria, we observe that the average gain is quite close to 0, for each target accuracy. 
The section 95\% is an exception, since its last three columns show a slowdown up to 9 rounds of communication, on average. 
This behavior is probably due to the adverse effects we already observed for the individual criteria in this dataset.

We can also observe the same behavior in Figure \ref{fig:iid}, where the curves of the combinations of criteria --- and the ones of the individual criteria --- are virtually indistinguishable.

\paragraph{MNIST with Non-IID distribution.}
Once we combine different criteria, we may observe several different behaviors.
If we focus on Table \ref{tbl:noniid}, we can observe two different trends. 
First, if we analyze the data in the table moving from the left to the right, we may observe an increase of negative gains in the sections related to 90\% and 95\% of the accuracy.
 Even here, if we focus on the last three columns, we may notice an average delay that can reach up to 12 rounds.
However, even in this case,  there are some combinations that achieve an advantage over DS: LD $\succ$ DS $\succ$ MW, MW $\succ$ DS $\succ$ LD, and MW $\succ$ LD $\succ$ DS.
LD $\succ$ DS $\succ$ MW shows an interesting trend since it reaches 70\%, 80\%, 90\%, 95\% of accuracy 2, 2, 0, 1 rounds sooner than DS, respectively.
Here, the advantage is higher for the lower accuracy thresholds, while it is almost 1 for 95\% accuracy.
Instead, MW $\succ$ DS $\succ$ LD shows a better and incremental trend.
Indeed, in this case, the advantage of rounds is -0.1, 0.2, 2.1, and 1.7, respectively.  
Finally, MW $\succ$ LD $\succ$ DS shows a bad advantage average for 70\%, and 80\% (0.2, and -0.6), while for 90\%, and 95\% it shows an interesting performance boost by reaching the goals 3.1, and 4.7 rounds before DS.

\paragraph{CelebA.}
Results in Table~\ref{tbl:celeba} show  results which are coherent with the effects of the individual criteria.
Rows DS $\succ$ IS $\succ$ CB, as well as IS $\succ$ DS $\succ$ CB, IS $\succ$ CB $\succ$ DS, and CB $\succ$ IS $\succ$ DS, denote the negative influence of the IS criterion. 
In particular, they show a little boost in very few cases for a low percentages of devices, but their averages generally show a slowdown from 2 up to 13 rounds of communication.
On the other side, we may observe the effects of BC, which is beneficial even when combined with the other criteria.
Even then, this effect is more evident when IS has the last priority (i.e., by minimizing its influence). 
In this case, we observe a substantial speed-up against the sole DS criterion, and also when compared with the CB individual criterion. 
In detail, permutation DS $\succ$ CB $\succ$ IS shows the best results for all target values of accuracy and all values of coverage for devices, up to 12 rounds of communications (against the sole DS criterion). 
Even Figure \ref{fig:celeba} clearly shows these behaviors.
Specifically, we may observe the cyan line (DS $\succ$ CB $\succ$ IS) against the red line (DS only) and the blue line (CB only). 

\subsection{Discussion}
\label{sec:discussion}
The analysis in the previous section already shows the fundamental role of the identification of local criteria.
By considering the underlying dataset, the different criteria heavily affect the training phase outcome.

As an example, when we create a federated dataset by randomly picking samples from an original dataset (i.e., with an IID distribution), the adoption of statistical criteria is not beneficial, since data, on local devices, shows the same expectation.

On the other hand, even in MNIST Non-IID distributed dataset, it is quite disappointing that we can not appreciate the benefits of introducing new criteria like label diversity.
However, if we observe the dataset deeper,  it contains local shards of the same size, and each client processes, at most, samples of two digits that make the number of classes in local datasets a piece of useless information. 

Instead, the criterion on model divergence gives an initial boost to training, but it seems to have adverse effects for higher target values of accuracy. 
Even here, the behavior is interesting and predictable.
Indeed, the penalization of significant divergences is the right choice because it helps to build a robust estimator on the average of the samples.
However, at the end of the training, higher precision and fine training could require those differences.

As expected, when we observe the most realistic dataset, CelebA, we may appreciate more the effects of the proposed approach.
In fact, it is a realistic Non-IID distribution of images where each client holds samples which are different in number and number of classes. 
Consequently, the attribute about class balancing gives the best results either when considered individually and also combined with the dataset size attribute. 
This is probably due to the differences both in dataset size and class balancing among the local datasets. 
Conversely, Sharpness is a theoretically useful attribute, but it results in a slowdown of performance during the training phase. 
We believe that the reason for this behavior is twofold.
First, only a few images are marked as blurry in the dataset.
Second, a right combination of sharp and blurry images has widely proven to be beneficial for the generalization of a CNN.

\section{Conclusions and Future Perspectives}
\label{sec:conlusion}
In this work, we have introduced a practical Federated Learning (FL) protocol that exploits non-sensitive client information to aggregate the local models. 
In detail, we have formalized the notion of a local criterion for clients in an FL scenario, and the notion of priority ranked criteria.
By considering the ranking of the criteria, we have defined the score functions to weigh the contribution of a client.
We have tested our approach on three well known federated learning datasets: IID-distributed MNIST, Non-IID-distributed MNIST, and CelebA.

The experiments show that we can substantially improve the standard federated learning approach by exploiting a properly defined set of local criteria. 
The approach is particularly effective when dealing with Non-IID distributions of data.
This is remarkable since a real federated scenario will show a Non-IID distribution of data, rather than an IID distribution, where differences among clients are not so perceptible.  In general, substantial differences among devices about a property make the corresponding criterion more effective.
In practical terms, we have improved the federated approach by enhancing the individual differences, respecting the true federated learning spirit. In the experiments, for each dataset, we have investigated three different criteria that consider local datasets and local models. 

However, there is still a broad spectrum of criteria that deserves to be explored. 
On the other hand, from the experiments, it also emerges that a criterion's efficacy highly depends on the specific dataset/domain.
Even though several considerations may lead the researcher, it is not possible to find a unique criterion that meets the needs of all the possible domains.
Nevertheless, the study of the individual criteria has revealed some information about their impact on the training phase, which we hope may be useful for other researchers.

Next, we have observed how some criteria may be useful in some moments of the training, but they also may cause issues in others. 
In this respect, we propose a self-adaptive model that re-ranks the priority of criteria as partially investigated in~\cite{DBLP:conf/aiia/AnelliDNF19}.
We are particularly interested in this research direction and in the chance of leveraging this federated approach to other machine learning scenarios, like Recommender Systems~\cite{ABDLTD17, ADDRT19a} and Deep Learning.
Indeed, in those research fields, local model specialization and privacy are crucial, and this prioritized federated learning approach may be particularly beneficial. 

Finally, currently the focus of our work has been to protect privacy of users and provide satisfactory predictions. In doing this, as our proposed system relies on a number of local criteria obtained from users, using such criteria may lead to some unfair or biased recommendation toward particular class of users (e.g., men v.s. women) or particular class of items (paid v.s. non-paid jobs). The reason is we do not control over how to use these features to protect user fairness and algorithmic biases. We deem this  an important open research direction for which we plan to employ fairness-aware metrics as part of our approach~\cite{DBLP:journals/jmlr/ZafarVGG19,DBLP:conf/recsys/DeldjooAZKN19}. 

\paragraph{Acknowledgements} The authors wish to thank Domenico Siciliani for helping with the implementation of the framework. The authors acknowledge partial support from Innonetwork Apollon, Innonetwork CONTACT, Exprivia Digital Future, PON OK-INSAID, PON FLET.   

\bibliographystyle{unsrtnat}
\bibliography{sample}

\end{document}